\documentclass{article}
\usepackage{enumitem}
\usepackage{mdframed}
\usepackage{pdfpages}
\usepackage{fullpage}
\usepackage{tabularx,booktabs}
\usepackage{wrapfig}
\usepackage[utf8]{inputenc} 
\usepackage[T1]{fontenc}    
\usepackage{hyperref}       
\usepackage{url}            
\usepackage{booktabs}       
\usepackage{amsfonts}       
\usepackage{nicefrac}       
\usepackage{microtype}      
\usepackage{lipsum}
\usepackage{fancybox}   
\usepackage[export]{adjustbox}
\usepackage{fancyhdr}       
\usepackage{graphicx}       
\graphicspath{{media/}}     
\usepackage{seqsplit}
\usepackage{pgffor} 
\usepackage{ulem}

\usepackage{tabularx}
\usepackage{enumitem}
\usepackage{listings}
\usepackage{xcolor}

\lstdefinelanguage{json}{
    basicstyle=\ttfamily\small,
    showstringspaces=false,
    breaklines=true,
    morestring=[b]",
    morecomment=[l]{//},
    stringstyle=\color{blue},
    commentstyle=\color{gray},
}

\usepackage{subcaption}
\usepackage{amsthm,amssymb,amsfonts}
\usepackage{tikz,lipsum,lmodern}
\usepackage[most]{tcolorbox}
\definecolor{cream}{RGB}{222,217,201}

\usepackage{xcolor}
\usepackage{listings}

\usepackage{caption}
\usepackage{PRIMEarxiv}
\usepackage{multirow}
\usepackage[utf8]{inputenc} 
\usepackage[T1]{fontenc}    
\usepackage{hyperref}       
\usepackage{url}            
\usepackage{booktabs}       
\usepackage{amsfonts}       
\usepackage{nicefrac}       
\usepackage{microtype}      
\usepackage{lipsum}
\usepackage{fancyhdr}       
\usepackage{graphicx}       
\graphicspath{{media/}}     

\usepackage{tabularx,booktabs}
\usepackage{wrapfig}
\usepackage[utf8]{inputenc} 
\usepackage[T1]{fontenc}    
\usepackage{hyperref}       
\usepackage{url}            
\usepackage{booktabs}       
\usepackage{amsfonts}       
\usepackage{nicefrac}       
\usepackage{microtype}      
\usepackage{lipsum}
\usepackage{fancyhdr}       
\usepackage{graphicx}       
\graphicspath{{media/}}     
\usepackage{seqsplit}
\usepackage{geometry}
\usepackage{caption} 

\usepackage{subcaption}
\usepackage{amsthm,amssymb,amsfonts}
\usepackage{tikz,lipsum,lmodern}
\usepackage[most]{tcolorbox}
\definecolor{cream}{RGB}{222,217,201}

\newtcolorbox{Box1}[2][]{
                lower separated=false,
                colback=white!80!gray,
colframe=black, fonttitle=\bfseries,
colbacktitle=black!50!gray,
coltitle=black,
enhanced,
attach boxed title to top left={xshift=0.5cm,yshift=-2mm},
title=#2,
boxrule=0.5pt,
boxed title style={colframe=black, boxrule=0.5pt},
#1}

\newtcolorbox{Box2}[2][]{
                lower separated=false,
                colback=white!80!white,
colframe=black, fonttitle=\bfseries,
colbacktitle=white!50!white,
coltitle=black,
enhanced,
attach boxed title to top center={yshift=-2mm},
title=#2,
boxrule=0.5pt,
boxed title style={colframe=black, boxrule=0.5pt},
#1,}

\hypersetup{
pdftitle={Sparks: Multi-Agent Artificial Intelligence Model Discovers Protein Design Principles},
pdfsubject={q-bio.NC, q-bio.QM},
pdfauthor={Markus J. Buehler, MIT, mbuehler@MIT.EDU},
pdfkeywords={Scientific Artificial Intelligence, Multi-agent system, Large Language Model, Materials Design, Scientific Discovery},
}
  
\title{Sparks: Multi-Agent Artificial Intelligence Model Discovers Protein Design Principles
\thanks{\textit{\underline{Citation}}: 
\textbf{A. Ghafarollahi, M.J. Buehler. arXiv, DOI:000000/11111., 2025}} 
}

\author{
  Alireza Ghafarollahi \\
  Laboratory for Atomistic and Molecular Mechanics (LAMM)\\Massachusetts Institute of Technology\\ 77 Massachusetts Ave.\\ Cambridge, MA 02139, USA 
   \And
  Markus J. Buehler \\
  Laboratory for Atomistic and Molecular Mechanics (LAMM)  \\
  Center for Computational Science and Engineering\\ Schwarzman College of Computing\\ Massachusetts Institute of Technology\\77 Massachusetts Ave.\\Cambridge, MA 02139, USA\\ \\
  Correspondence: \texttt{mbuehler@MIT.EDU} \\
}

\begin{document}
\maketitle
\begin{abstract}
Advances in artificial intelligence (AI) promise autonomous discovery, yet most systems still resurface knowledge latent in their training data. We present Sparks, a multi-modal multi-agent AI model that executes the entire discovery cycle that includes hypothesis generation, experiment design and iterative refinement to develop generalizable principles and a report without human intervention. Applied to protein science, Sparks uncovered two previously unknown phenomena: (i) a length-dependent mechanical crossover whereby beta-sheet-biased peptides surpass alpha-helical ones in unfolding force beyond ~80 residues, establishing a new design principle for peptide mechanics; and (ii) a chain-length/secondary-structure stability map revealing unexpectedly robust beta-sheet-rich architectures and a ``frustration zone'' of high variance in mixed alpha/beta folds. These findings emerged from fully self-directed reasoning cycles that combined generative sequence design, high-accuracy structure prediction and physics-aware property models, with paired generation-and-reflection agents enforcing self-correction and reproducibility. The key result is that  Sparks can independently conduct rigorous scientific inquiry and identify previously unknown scientific principles.
\end{abstract}

\keywords{Scientific Artificial Intelligence \and Multi-agent system \and Multi-modal intelligence \and Large language models \and Materials design \and Scientific Discovery \and Foundation models}

\section{Introduction}

From Newton’s laws to the discovery of DNA, science has long depended on human intuition, trial-and-error, and incremental experimentation. The 20th century saw the rise of computational modeling and automated data analysis, accelerating discovery but keeping humans at the core of the process. The advent of deep learning in the 2010s enabled machines to recognize complex patterns, but these systems remained tethered to their training distributions~\cite{NIPS2012_c399862d,10.1162/neco.1997.9.8.1735,kuhn1962structure,6793380,reddy2024towards}.
Most contemporary AI systems excel at statistical generalization within their training distribution, but they rarely generate or validate hypotheses that reach beyond it. Scientific discovery, however, demands more than pattern recognition; it requires agency of competing interests that can propose, test, and revise ideas until a falsifiable, general law emerges~\cite{kuhn1962structure,6793380,reddy2024towards,doi:10.1126/science.1165620,fodor1988connectionism,doi:10.1126/science.1165893,Buehler2025GA,Buehler2025SOG,Buehler2024PRefLexOR,Buehler2025GPReFLexOR}. We show that this capability arises when large foundation models are organized into an adversarial, task-specialized generation–reflection architecture: each proposer—charged with formulating a hypothesis, writing code, collecting new physics data from simulation, or interpreting data—is paired with an isomorphic critic that immediately interrogates the output, driving exploration into regions where their priors diverge. This adversarial loop pushes the search beyond the models' training distribution and enables the synthesis of genuinely novel knowledge. We instantiate the concept in protein science, an arena where the combinatorial space of sequences, structures, and mechanics has long defied exhaustive human exploration~\cite{dill2012protein,kuhlman2019advances,keten2008geometric, keten2010nanoconfinement}.

\begin{figure}[ht!]
\centering
    \includegraphics[width=1\textwidth]{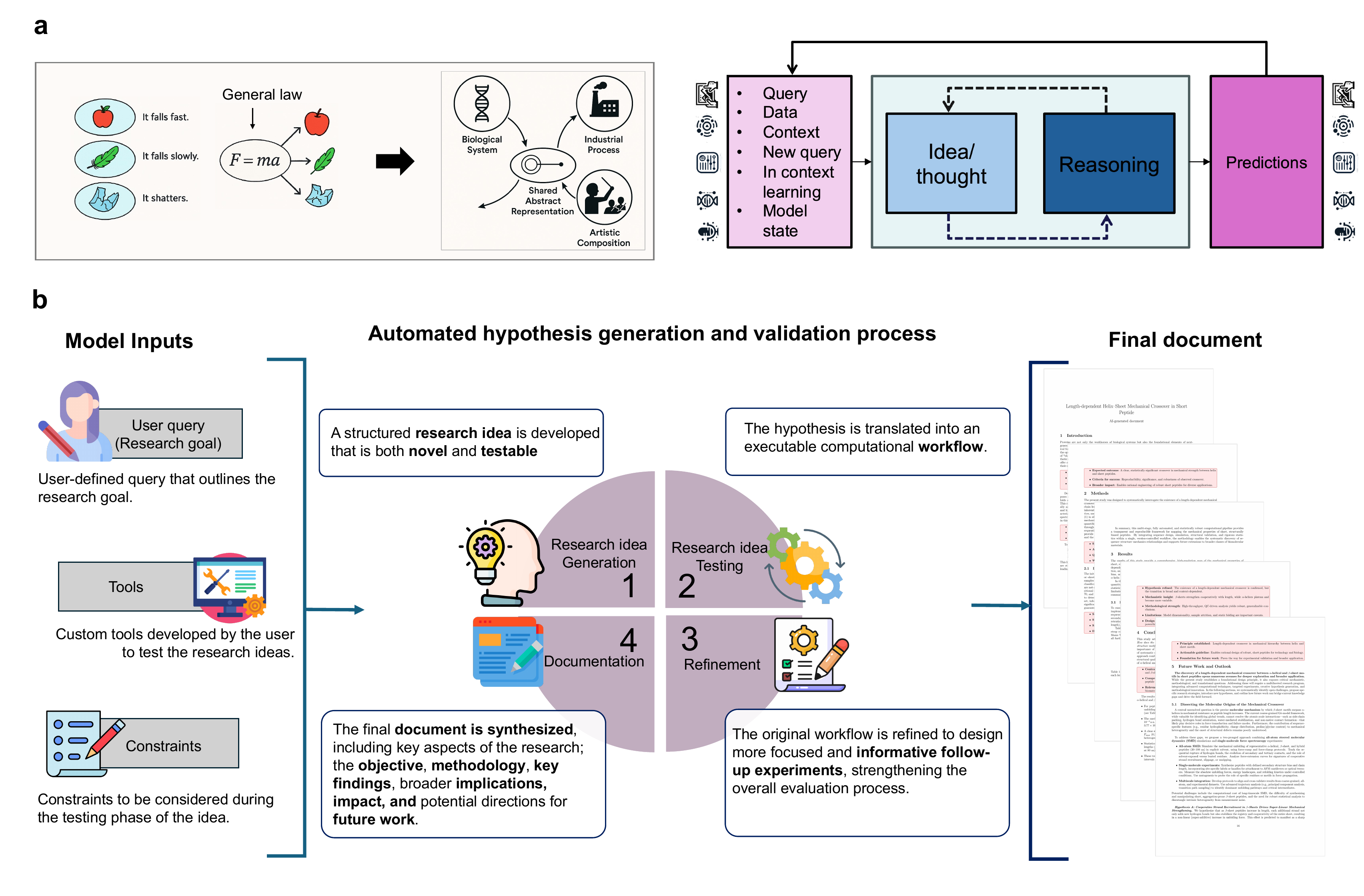}
\caption{
\textbf{Overview of Sparks, a multi-agent AI model for automated scientific discovery.} Panel a: Contemporary AI systems excel at statistical generalization within known domains, but rarely generate or validate hypotheses that extend beyond prior data, and cannot typically identify shared principles across distinct phenomena. This is because powerful models tend to memorize physics without discovering shared concepts. For scientific discovery, however, the elucidation of more general and shared foundational concepts (such as a scaling law, design principle, or crossover) is critical, in order to create significantly higher extrapolation capacity. Panel b: Sparks automates the end-to-end scientific process through four interconnected modules: 1) hypothesis generation, 2) testing, 3) refinement, and 4) documentation. The system begins with a user-defined query, which includes research goals, tools to test the hypothesis, and experimental constraints to guide the experimentation. It then formulates an innovative research idea with a testable hypothesis, followed by rigorous experimentation and refinement cycles. All findings are synthesized into a final document that captures the research objective, methodology, results, and directions for future work, in addition to a shared principle (such as in the examples presented here a scaling law or mechanistic rule). Each module is operated by specialized AI agents with clearly defined, synergistic roles.
}
    \label{fig:overview}
\end{figure}

Recent advances in artificial intelligence (AI) have begun to transform the field of protein design. AI-driven breakthroughs have fundamentally altered what is possible in protein science~\cite{huang2016coming, kortemme2024novo}. The advent of AlphaFold, for example, demonstrated that deep learning models can predict protein structures with unprecedented accuracy, solving a problem that had eluded researchers for decades \cite{jumper2021highly}. In parallel, generative models such as Chroma have enabled \textit{de novo} protein engineering, facilitating the creation of novel folds and functions beyond those found in nature \cite{ingraham2023illuminating}. Deep learning-based surrogate models, capable of predicting protein properties from sequence or structure or vice-versa, have further empowered researchers to optimize protein function at scale \cite{khare2022discovering, ni2023generative,ni2024forcegen}. However, these models are limited when it comes to integrating knowledge beyond their training domain—a capability that is essential for genuine scientific discovery, which relies on an iterative deep reasoning process.

In recent years, large‐scale foundation models, typified by OpenAI’s \texttt{GPT-4o}, \texttt{o1} and \texttt{o3} or Google \texttt{Gemini}, have become a powerful paradigm to aid in scientific research and other complex tasks~\cite{OpenAI2023GPT-4Report,openai2024gpt4ocard,openai2024openaio1card,bubeck2023sparksartificialgeneralintelligence}. Pre-trained on vast multimodal corpora of text, images, audio and domain-specific data, these reasoning engines can perform cross-domain inference, articulate context-aware hypotheses and generate human-readable scientific narratives \cite{vaswani2017attention,wei2022emergent,huang2022towards,chang2024survey}. When embedded in a multi-agent framework, individual models take on specialized roles and coordinate via structured messages, enabling collaborative intelligence with in-situ learning \cite{guo2024large}. Coupled to domain tools, simulation engines and knowledge graphs, such systems evolve into autonomous research platforms that generate novel ideas, automate complex materials-design tasks and accelerate discovery—exemplified by SciAgents and related efforts \cite{ghafarollahi2024sciagents,ghafarollahi2024protagents,jablonka2024leveraging,zheng2025large,ghafarollahi2025automating}. Other incipient multi-agent systems have been applied to automate research workflow and assisting scientists across various stages of the scientific process across diverse fields \cite{boiko2023autonomous, m2024augmenting, lu2024ai, schmidgall2025agent, si2024can, bran2023chemcrowaugmentinglargelanguagemodels}. For example, ``The AI Scientist'' was proposed~\cite{lu2024ai} as a fully autonomous system capable of executing the entire scientific pipeline for AI research. More recently, Google introduced the AI Co-Scientist, a multi-agent system designed for ``scientist-in-the-loop'' collaboration, positioning AI as an intelligent assistant that can augment human researchers \cite{gottweis2025towards}. However, earlier systems primarily targeted discovery within the field of artificial intelligence itself. In physics and related scientific domains, most prior work has focused on hypothesis generation without physics-informed validation, or has left experiment design and validation outside the autonomous core, relying instead on human intervention or external tools.

Despite these advances, the challenge of autonomous discovery in the physical sciences, such as materials science or protein science, remains formidable. Achieving independent, original discovery demands far more than the ability to generate plausible hypotheses; it requires intelligent systems that can integrate diverse forms of knowledge, combine data-driven and physics-based modeling for accurate predictions, and iteratively coordinate multiple dependent steps toward a solution, specifically targeting a novel discovery with generalizable principles as an outcome. Crucially, such systems must design and execute experiments, interpret results, and refine their strategies as new findings emerge. The end goal is not only to automate routine research tasks, but to meet the scientific standards of novelty, reproducibility, and physical meaningfulness—criteria that require both methodological rigor and creative intuition. As we discuss in this study, realizing this vision calls for a new generation of multi-agent systems powered by advanced reasoning AI agents, capable of seamlessly integrating tools, logic, and reasoning in a deeply nonlinear, feedback-driven workflow that can advance the frontier of autonomous scientific discovery.

In this study, we present Sparks, a multi-modal, multi-agent AI model that demonstrates scientific discovery by autonomously orchestrating the entire research process. By combining the reasoning and planning abilities of advanced foundation models with powerful domain-specific computational tools, Sparks enables specialized agents to collaborate  seamlessly through key stages of research. This system generates original hypotheses grounded in scientific reasoning, designs and implements computational experiments, interprets results, and adapts its research strategy in real time, without human intervention. Leveraging this integrated approach, Sparks independently proposed a novel hypothesis and then validated it through an iterative experimental approach, leading to a previously unreported, length-dependent mechanical crossover in short protein peptides, establishing a new design principle with broad implications. 

Our study marks a significant advance in autonomous scientific discovery by showing that an AI system can transcend its prior knowledge and independently uncover two previously unknown phenomena in protein science. First, Sparks discovered a length-dependent mechanical crossover: beta-sheet–biased peptides surpass their alpha-helical counterparts in maximal unfolding force once the chain length exceeds ~80 residues, establishing a new design principle for peptide mechanics. Second, the framework mapped a chain-length by secondary-structure stability landscape, revealing an unexpected intrinsic robustness of beta-sheet-rich proteins and a pronounced ``frustration zone'' of high conformational variance for mixed alpha/beta folds at moderate lengths. These breakthroughs were achieved with Sparks, our multi-modal, multi-agent AI framework purpose-built for hypothesis-driven research, which autonomously generated hypotheses, designed and executed simulations, iteratively refined its approach, and synthesized the findings without human intervention. Our results establish that AI can independently conduct rigorous scientific inquiry and deliver previously unknown discoveries.

The remainder of this paper is organized as follows. Section \ref{sec: Sparks} introduces Sparks, detailing its agentic architecture, agent roles, and the iterative workflow that underpins automated scientific discovery. Section \ref{sec: discussion} discusses the broader implications of this work, current limitations, and future directions for AI-driven scientific discovery. We provide a detailed analysis of the scientific work done by the AI model and provide the full traceable output for two examples as Supplementary Materials.

\section{Results}\label{sec: Sparks}
We first introduce Sparks and its foundational design concepts, as a tool for driving the scientific discovery process. We then present two case studies to demonstrate how this intelligent system autonomously conducts research by integrating the synergistic capabilities of specialized AI agents and computational tools.  

\subsection{Multi-agent model for automated hypothesis-driven scientific discovery}

Figure~\ref{fig:overview} illustrates the outline of our proposed multi-agent system designed to automate the scientific discovery process. The inputs to the model are a scientific query submitted by the user, a list of available tools, and imposed experimental constraints. The model consists of four main modules which are executed sequentially following a predefined sequence of tasks that ensures consistency and reliability through the scientific discovery process; 1) Idea generation, 2) idea testing, 3) refinement, and 4) documentation. This high-level flow captures the progression from abstract questions to concrete hypotheses, from idea implementation and testing to evaluation and refinement, and ultimately to the synthesis of findings and documentation. Sparks is a multi-agent model where the scientific discovery pipeline is by agents carefully designed via prompting to play a specific role. 

\begin{figure}[ht!]
\centering
    \includegraphics[width=1\textwidth]{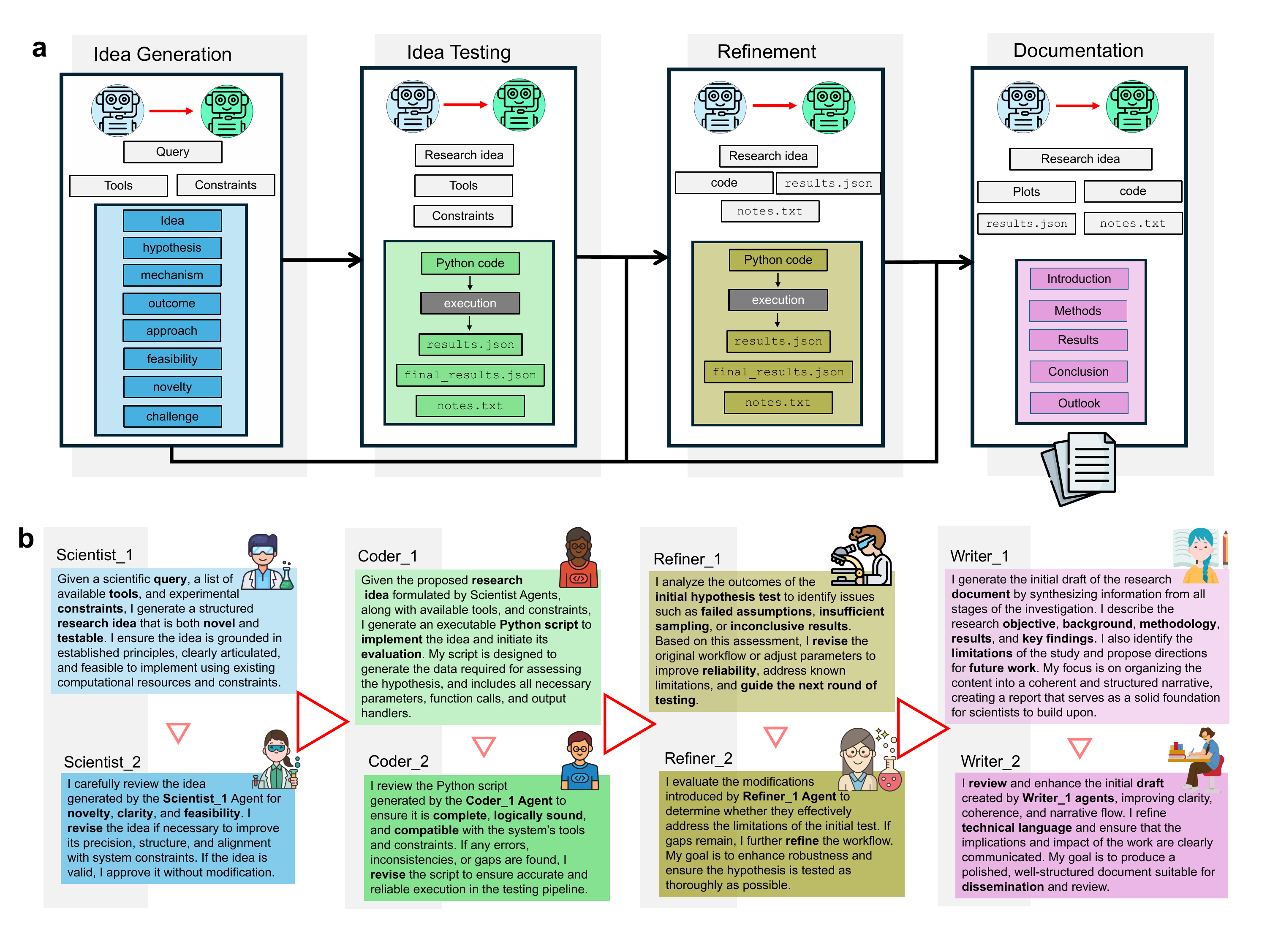}
\caption{
\textbf{(a) Overview of the entire process from idea generation to the final document.} First, the Idea Generation module formulates a high-impact research idea. Then in the testing module translates these hypotheses into executable workflows, autonomously conducting simulations or analyses to generate quantitative results. The refinement module is responsible for refining the testing strategy based on the results, adaptively revising the experimental design through an iterative feedback loop that sharpens insight and prompts reliable hypothesis testing; and writer agents consolidate the entire research lifecycle into a comprehensive document that not only presents key findings and methodologies, but also outlines future research directions—effectively serving as a blueprint for subsequent scientific inquiry.
\textbf{(b) Overview of the AI Agents and their role implemented in Sparks.}
Each module operates through a structured yet adaptive sequence of agent interactions, enabling consistency and context-aware responses across the research workflow. Each agent dynamically adapts to previous content in real time, ensuring. Inter-modular agents facilitate a generation–reflection strategy, using dynamic prompts to process evolving inputs and coordinate outputs, ensuring the system adapts fluidly to new insights throughout the research process.}
    \label{fig:overview_2}
\end{figure}

An overview of the entire automated AI-driven process is illustrated in Figure \ref{fig:overview_2}(a), from idea generation to the final document. The process begins with the Idea Generation module which, in response to the User's query and inputs, generates a novel research hypothesis that is both novel and feasible within the system’s computational constraints. Following hypothesis generation, the system enters the Idea Testing module, where the hypothesis is tested. This phase includes writing a Python script that carefully implements the idea by calling and executing the tools, processing and storing outputs in structured JSON format, and documenting the workflow for explainability and reproducibility. Next, the results are provided to the Refinement module for evaluation and assess whether the data collected sufficiently address the research question. If not, the system autonomously designs and performs additional follow-up experiments. This loop continues until convergence or until the user-defined testing limit or experiment constraints are reached. This mechanism ensures that the system neither under-explores nor expends unnecessary computational resources, adapting its workflow based on observed outcomes. Finally, Sparks transitions to the documentation phase where the system integrates all the previous content and results and writes a comprehensive report. In this phase, the system synthesizes the entire research trajectory, generating visualizations, summarizing findings, and compiling them into a structured scientific document. The report is designed to be interpretable, verifiable, and reusable, supporting external human analysis. 

Each module in Sparks is empowered by a set of AI agents with specialized prompting to yield a reliable, complete, and accurate response. Each agent has a profile describing its role in the system. An overview of the Agents recruited in Sparks is shown in Figure \ref{fig:overview_2}(b) aligned with a description of their role in the system. Certain agents also have the ability to solicit new physical data and insights, such as by setting up and running molecular dynamics simulations.

In the following sections, we provide an in-depth exploration of each module in Sparks and the intermodular agents. Together, they form a system that does more than predict or generate outputs, it engages in a sustained, reflective scientific workflow. By combining hypothesis-driven exploration, computational implementation, and structured iteration, Sparks represents a step toward systems that do not merely assist science but actively practice it.

\subsection{Idea Generation: formulating novel and feasible idea}
The idea generation module serves as a foundational component of the system, initiating the scientific discovery process through the creation of structured, testable research ideas. This module is composed of two specialized agents, \texttt{scientist\_1} and \texttt{scientist\_2}, each guided by carefully designed prompts that define their respective roles and responsibilities.

The process begins with a query, which serves as the entry point for the discovery workflow. The query represents a high-level scientific task or research goal posed by the user. It defines the overall direction of the investigation and serves as the conceptual anchor from which all hypotheses are derived. The query may be broad or targeted but must be interpretable within the system's constraints and executable using the available tools.

\begin{figure}[ht!]
\centering
    \includegraphics[width=1\textwidth]{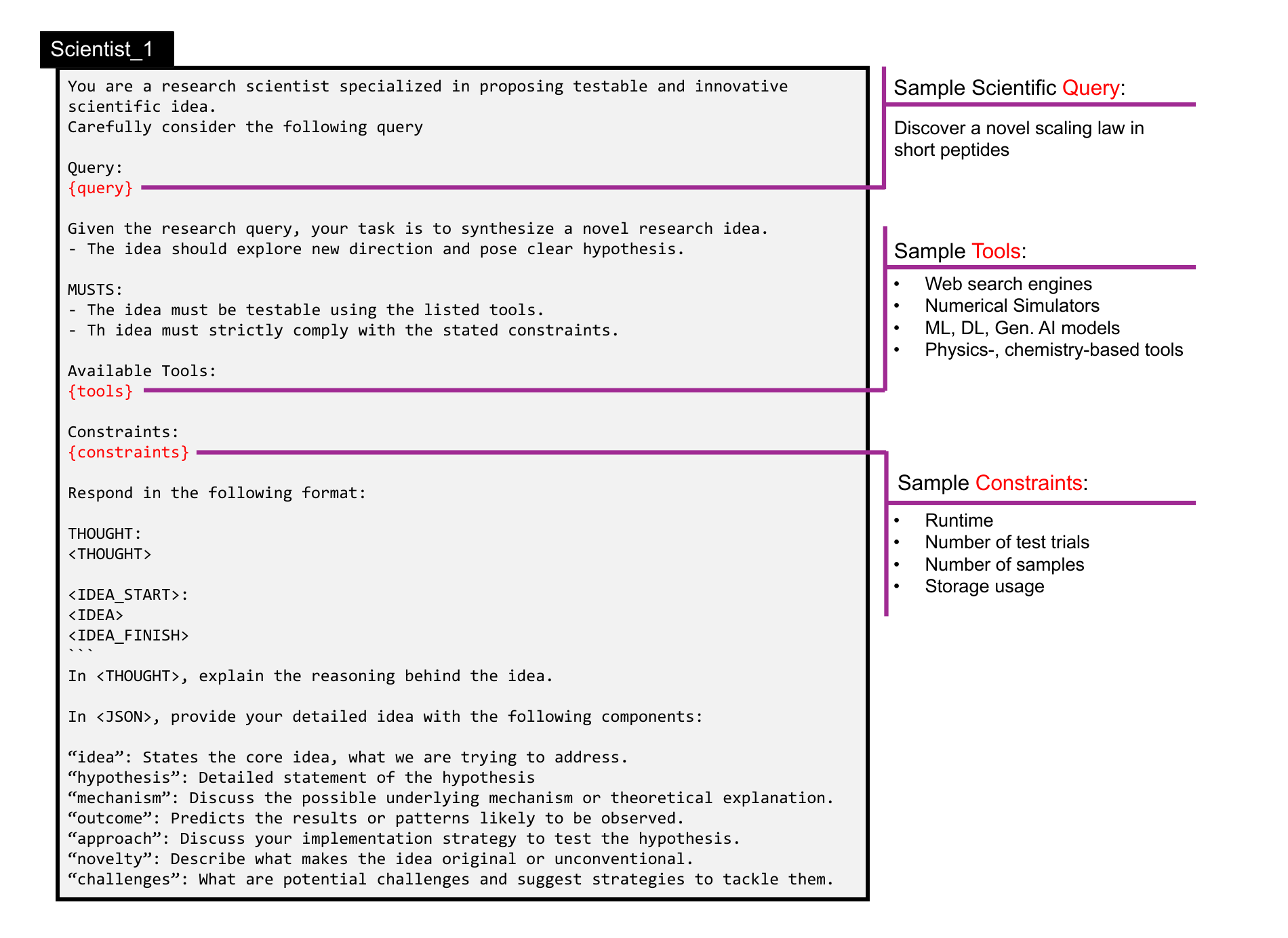}
\caption{\textbf{Overview of the \texttt{Scientist\_1} prompt}. The prompt takes the user's query, a list of available tools, and experiment constraints as input. The agent is tasked with generating novel research ideas, guided by instructions to promote both novelty and feasibility. The agent is required to return a research idea containing several key component.}
    \label{fig:scientist_1}
\end{figure}

\begin{itemize}
    \item \texttt{\textbf{Scientist\_1}}:
Given the query, tools, and constraints, through careful prompting as shown in Figure \ref{fig:scientist_1}, the scientist\_1 is tasked with formulating a novel, scientifically sound, and computationally feasible research idea. The agent receives a structured prompt containing the query, a list of accessible tools, and a set of experimentation constraints. The output of this agent is a structured research proposal composed of the key components: idea, hypothesis, mechanism, outcome, approach, feasibility, novelty, and challenge.
\item \texttt{\textbf{Scientist\_2}}
After \texttt{scientist\_1} submits the initial proposal, the \texttt{scientist\_2} agent reviews the idea and reflects on its clarity, novelty, feasibility, and alignment with system constraints. This agent operates using a structured prompt that mirrors the format of the original idea, allowing it to directly assess and revise each component if necessary. The reflection process may include improving the scientific precision of definitions, refining the hypothesis structure, or modifying the approach to better fit the system's operational limits. If the proposed idea meets all criteria, the reflection agent returns it unmodified. Otherwise, it produces a revised version, ensuring that only well-formed and executable hypotheses proceed to the testing phase.
\end{itemize}

Together, the generation (\texttt{scientist\_1}) and reflection (\texttt{scientist\_2}) agents form a tightly coupled mechanism for idea formulation. Their interaction ensures that research ideas generated by the system are not only creative and exploratory but also practical, testable, and fully aligned with the computational capabilities of Sparks. This module thus serves as a critical foundation for iterative, automated scientific exploration.

\subsection{Idea Testing: Translating hypothesis into executable workflows}\label{sec:testing}
This module serves as another core component of the scientific reasoning and validation process, transforming abstract research ideas into concrete, testable investigations. It is responsible for implementing the proposed approach and generating evidence to evaluate the underlying hypothesis. Guided by the defined objectives, available tools, and experimental constraints, the system constructs an executable workflow—typically in Python—to simulate and analyze the hypothesis in a systematic manner. The goal is to enable rigorous, transparent, and repeatable testing within the boundaries of the system’s capabilities.

This module focuses on a first-pass implementation of the idea, aiming to validate its feasibility under current conditions, and is managed by two agents that work in tandem to construct the computational workflow  

\begin{itemize}
    \item \texttt{\textbf{Coder\_1}}:
The coder agent is responsible for generating a Python script that faithfully implements the research idea. The agent receives a structured prompt that includes the original query, the hypothesis to be tested, a list of available computational tools, and a set of implementation constraints. The script must be self-contained and executable, and all results must be stored using a standardized format to support reproducibility and downstream analysis. Specifically, raw and intermediate outputs are saved in results.json, final outcomes in final\_results.json, and workflow metadata—including parameter choices, file names, function usage, and relevant justifications—are recorded in notes.txt.

\item \texttt{\textbf{Coder\_2}}:
Once the script is generated, the coder reflection agent evaluates it for technical correctness, completeness, and adherence to system standards. This agent checks whether the appropriate tools have been used, whether the outputs are generated and saved properly, and whether the computational procedure aligns with the original hypothesis. If any issues are detected—such as omissions, incorrect function calls, or data-saving inconsistencies—the reflection agent modifies the code accordingly. If the script is deemed valid, it is passed forward without modification.

\end{itemize}

The division of responsibilities between these two agents improves reliability by ensuring that every hypothesis is translated into a consistent, reproducible, and verifiable computational procedure. All outputs are stored in a structured format, enabling traceability throughout the research process and supporting the next stages of discovery, where previous results may be revisited or extended as outlined below.

\subsection{Refinement: informative follow-up testing}
The refinement module plays a critical role in Sparks’s autonomous scientific workflow by serving as an iterative checkpoint after initial testing. Its primary goal is to reassess the hypothesis in light of early results and determine whether further experimentation is needed. Since initial tests may not fully capture the complexity of the scientific problem, this module ensures that conclusions are not drawn from incomplete data, failed assumptions, or inconclusive evidence. It is particularly valuable in cases where results exhibit high variability or where subtle trends warrant deeper investigation.

To perform this task, Sparks deploys two specialized agents in the refinement module:

\begin{itemize} \item \texttt{\textbf{Refiner\_1:}}
This agent is responsible for the first layer of evaluation. It is given full access to the contextual information from the previous round, including the hypothesis, implementation code, summary results (\texttt{final\_results.json}), detailed logs and raw outputs (\texttt{results.json}), and experiment notes (\texttt{notes.txt}). Refiner\_1 analyzes these files to assess the robustness and coverage of the initial testing. If gaps are identified—such as insufficient sampling, invalid assumptions, or ambiguous results—it generates a revised version of the experiment script. The updated code is designed to build on the previous run, reusing existing data and appending new results in a consistent manner. Care is taken to preserve variable naming conventions and file structures to ensure continuity and prevent data overwriting.

\item \texttt{\textbf{Refiner\_2:}}  
This agent reviews the decision and implementation proposed by \texttt{Refiner\_1}. If additional testing is deemed necessary, \texttt{Refiner\_2} validates the revised code for correctness, completeness, and internal consistency. It may also fix errors or enhance the code for better performance and reliability. If no follow-up is required, the agent terminates the testing loop and returns a definitive \texttt{"NO\_FOLLOWUP"} flag, signaling the system to proceed to the documentation phase.

\end{itemize}

This refinement loop can continue for multiple rounds (requiring increasing amounts of compute), enabling Sparks to iteratively improve its understanding of the problem. The process concludes when either no further testing is required or the system reaches a predefined limit on testing rounds, denoted by $N_{\text{test}}$. At that point, the workflow transitions to the documentation phase, where the results and insights are formally compiled.

\subsection{Documentation to Yield a Traceble, Coherent and Structured Narrative}

Once hypothesis generation and validation are complete, Sparks transitions to the documentation phase, where a detailed and structured research manuscript is generated. This document not only compiles results from previous stages but also refines the original proposal, offering enriched analysis and deeper insights. The manuscript includes plots, tables, and structured interpretations to enhance clarity, transparency, and scientific communication. Key elements such as insights, impacts, limitations, and future research directions are also addressed.

To ensure comprehensive and high-quality output, we implement a multi-stage documentation pipeline driven by specialized agents. Each agent is responsible for generating one specific section of the manuscript, Introduction, Methods, Results, Conclusion, and Outlook, using carefully designed prompting strategies. The modularity of this pipeline guarantees that each section is internally coherent and collectively forms a scientifically rigorous narrative.

Before the main documentation process begins, a dedicated plotting submodule is activated to visualize the results and enable multimodal interpretation. This submodule is driven by the following agents:

\begin{itemize} \item \texttt{\textbf{Plot\_Designer\_1:}}
This agent receives the research idea and results from preceding modules, selects the most suitable plot types to represent the data, and generates a Python script to create high-quality visualizations. It ensures that the plots reveal patterns, trends, or anomalies critical to hypothesis evaluation. The agent is instructed to, where applicable, perform regression analysis, overlay fitting curves, and return the associated fitting parameters in json format.

\item \texttt{\textbf{Plot\_Designer\_2:} } 
This agent reviews the code generated by Plot\_Designer\_1. It checks for accuracy, completeness, and clarity, and suggests improvements or additional visualizations to better support data interpretation. It may also revise the code to correct errors or enhance scientific utility.

\item \texttt{\textbf{Plot\_Analyzer:}}  
Once the plots are generated, this agent analyzes each figure in the context of the original hypothesis and research question. It provides a structured interpretation, describing observed patterns, correlations, and their implications. Additionally, it produces a caption for each figure and summarizes key insights in a concise format.

\end{itemize}

Following plot generation and analysis, the system proceeds to compile the manuscript written in \LaTeX. This stage is orchestrated by a suite of foundation model based writing agents:

\paragraph{Writing Agents} Each section of the manuscript is composed by a dedicated agent:

\begin{itemize} \item \texttt{\textbf{Introduction:}}
Presents the research objective, motivation, hypothesis, and challenges. It situates the problem within the broader scientific context.

\item \texttt{\textbf{Methods:}}  
Describes the design rationale, parameter space, computational workflow, and tools used. It also highlights the challenges encountered during implementation.

\item \texttt{\textbf{Results:}}  
Discusses key findings, interprets the outcomes, evaluates the hypothesis, and identifies scientific contributions and limitations.

\item \texttt{\textbf{Conclusion:}}
Summarizes the most important results, synthesizes insights, and articulates the study's broader significance.

\item \texttt{\textbf{Outlook:}}  
Identifies unresolved questions and proposes future research directions. It also suggests new hypotheses inspired by the current findings.

\end{itemize}

\paragraph{Reflection Agent} Each of the five writer agents is paired with a Reflection Agent responsible for reviewing the corresponding section. It checks for completeness, accuracy, depth, and clarity. The Reflection Agent may expand the text, clarify ambiguities, and enhance scientific detail. For each section, it also generates a highlighted summary box that distills the key takeaways—main findings, conclusions, and their implications.

Together, these agents collaboratively generate a final document that encapsulates the entire research process—from hypothesis and experimentation to interpretation and projection. This document serves as a springboard for human researchers to build upon the AI's work, refine methods, or explore new directions, facilitating a collaborative and iterative model of scientific discovery.

\subsection{Science Discovery Example I:  Length-dependent Helix–Sheet Mechanical Crossover in Short Peptides}\label{sec:example_1}

Sparks is designed as a domain-adaptable system for automated scientific discovery across a wide range of disciplines. Its flexibility stems from a modular architecture, where domain-specific tools, constraints, and instructions are supplied as inputs to adapt the system’s behavior to the research context. These inputs define the overall research objective, computational capabilities, and experimental limitations, guiding the scientific inquiry in a targeted, structured, and interpretable manner. Importantly, the agent prompts—including their objectives, operational instructions, and expected output formats—are defined in a domain-agnostic way, enabling Sparks to generalize scientific discovery workflows across disciplines.

In this study, we demonstrate Sparks’s capabilities in the context of protein design. Progress in this field is hindered by the vastness of amino acid sequence space, the intricacies of atomic-scale folding, and the complex mapping between structure and function. These challenges have historically made protein science highly dependent on expert intuition and trial-and-error approaches.

Sparks fills this gap by unifying hypothesis generation, computational planning, tool-based execution, and iterative reflection into a single AI-driven multi-agent framework. In this work, we showcase Sparks’s ability to autonomously formulate and test hypotheses, ultimately uncovering novel physical insights in short protein peptides. Notably, the system independently discovered a previously unreported size-dependent phenomenon—revealing a mechanical crossover effect in peptides of varying lengths.

\begin{figure}[ht!]
\centering
    \includegraphics[width=1\textwidth]{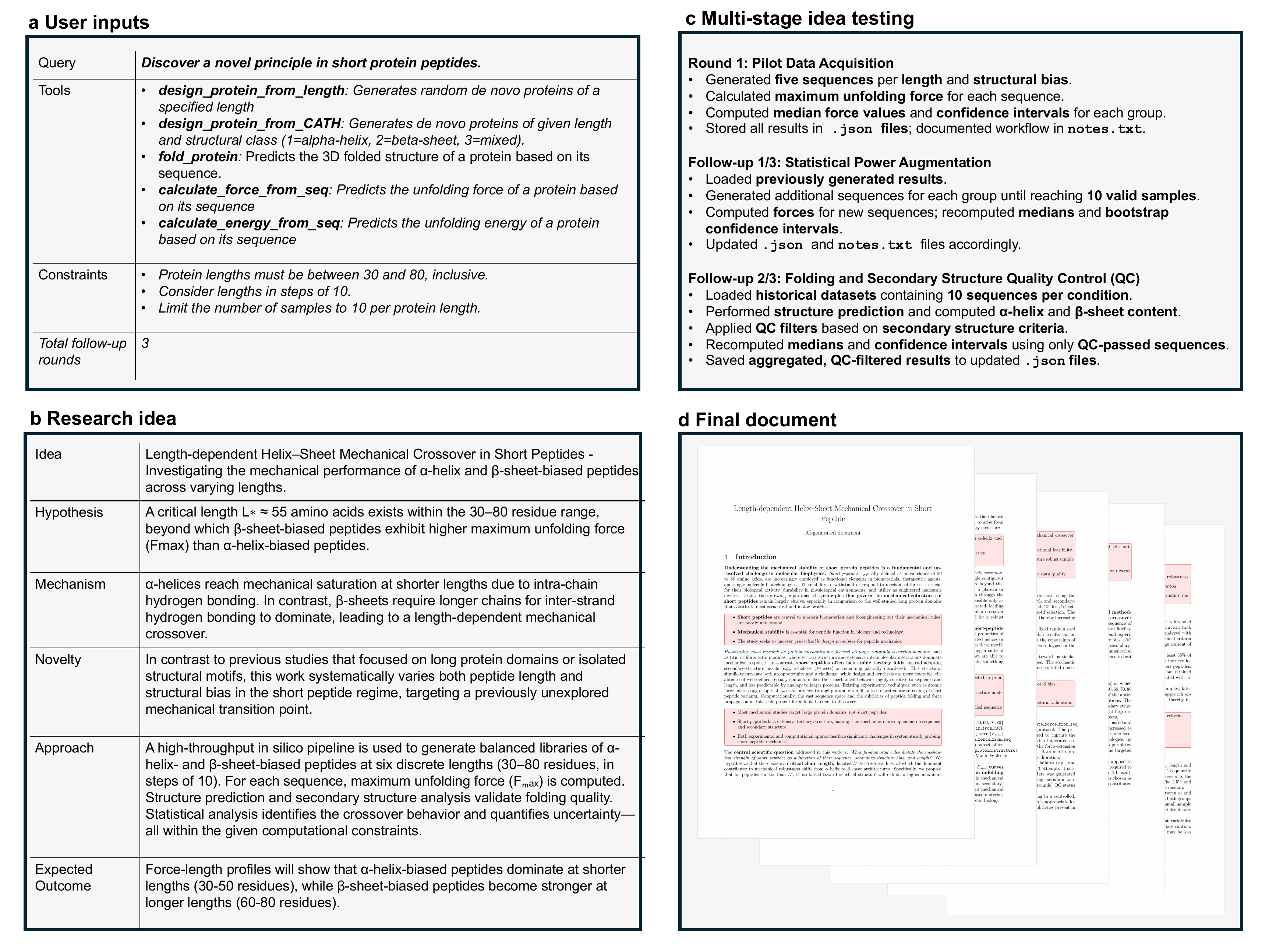}
    \caption{\textbf{Length-dependent helix–sheet mechanical crossover in short peptides.}  
    (a) User-submitted input describing the initial research query, accessible tools, experiment constraints, and total number of follow-up rounds $N_{\text{test}}$.  
    (b) Structured research idea generated by the AI model in the Idea Generation module.  
    (c) Multi-stage AI-driven evaluation of the hypothesis, incorporating testing and refinement.  
    (d) Final document created by the model, provides a comprehensive overview of the research, including key results that demonstrate the length-dependent mechanical crossover—where $\beta$-sheet-biased peptides surpass $\alpha$-helix-biased peptides in unfolding force as the peptide length increases. The full document is provided in Section \ref{sec: S1} of the SI.}
    \label{fig:model_example}
\end{figure}

The inputs to the system for this example are outlined in Figure \ref{fig:model_example}(a). To test the model’s capacity for creative exploration, we provide a general research query without directing the system toward any specific hypothesis space. Sparks is then equipped with a set of domain-relevant tools as shown in Figure \ref{fig:model_example}(a): generative models for creating structure-biased protein sequences of a given length, a folding tool to predict 3D structures and produce PDB files, and pre-trained autoregressive models to estimate mechanical unfolding forces and energies from sequences. For simplicity and efficiency, we define a small set of computational constraints. We also set the total number of follow-up rounds to 3. The complete research document autonomously generated by Sparks for this example is presented in Section \ref{sec: S1} in SI. In the following sections, we examine the results and the scientific insights derived from this fully automated investigation.

\paragraph{Ideation}
In response to the query, the scientist agents propose a research idea aimed at investigating a length-dependent mechanical phenomenon in short peptides. The central hypothesis posits the existence of a critical peptide length at which $\beta$-sheet-biased sequences exhibit greater mechanical strength than their $\alpha$-helical counterparts. The core components of this AI-generated research idea are summarized in Figure \ref{fig:model_example}(b).

As demonstrated, the AI-generated idea and hypothesis highlight the proficiency of Sparks in formulating well-defined, testable scientific hypotheses grounded in the principles of biomolecular mechanics. The hypothesis introduces a specific crossover length at which $\beta$-sheet-biased peptides are predicted to surpass $\alpha$-helical ones in mechanical strength—a nontrivial, length-dependent phenomenon. It also contextualizes the novelty of the idea through comparison with prior studies, highlighting a gap in existing literature. The model not only articulates the hypothesis clearly but also outlines a rigorous, computationally tractable approach for testing it. By proposing a systematic in silico pipeline that generates balanced peptide libraries, ensures structural fidelity through folding validation, and applies statistical analysis to quantify crossover trends and uncertainties, the model defines a complete experimental plan. This reflects its capacity to autonomously generate scientifically meaningful, mechanistically interpretable, and experimentally actionable hypotheses—thereby accelerating the path from idea to investigation and discovery in protein science.

\paragraph{Testing}
Next, the system proceeds with testing the hypothesized length-dependent mechanical crossover between $\alpha$-helix- and $\beta$-sheet-biased short peptides by implementing a structured and detailed evaluation strategy. As outlined in Section \ref{sec:testing}, the testing module operates in two phases: an initial implementation phase focused on pipeline validation and rapid hypothesis evaluation, followed by targeted follow-up rounds designed to address limitations identified in the earlier stage. 

Key elements of the experimental workflow executed in this example are summarized in \ref{fig:model_example}(c). The results of this iterative testing framework demonstrate the effectiveness and scientific rigor of the Sparks system in performing autonomous hypothesis evaluation:

\begin{itemize} \item The multi-round workflow enables incremental improvements in data quality, statistical power, and structural validation through systematic refinement. \item Each round builds upon the previous, supporting staged error detection, quality control, and data enrichment in a modular, extensible fashion. \item All data and metadata—including per-sample results, designed sequences, PDB structures, and Python scripts—are automatically archived, ensuring full traceability, transparency, and reproducibility. \item The entire pipeline—from hypothesis formulation to dataset generation, structure prediction, and statistical analysis—is executed end-to-end without human intervention. \end{itemize}

Hypothesis evaluation and testing is a critical step in the scientific discovery process, demanding structured workflows, iterative execution, and robust data analysis. Sparks addresses these needs through its multi-step testing module, offering a transparent, reproducible, and fully autonomous computational framework for scientific hypothesis testing.

\paragraph{Results}

We present the results produced by the AI model, as analyzed and structured by the plotting and documentation modules. The full analyses are available in Section \ref{sec: S1} of the Supplementary Information. Below, we summarize the main scientific contributions generated by Sparks, highlighting its effectiveness in uncovering novel scientific principles.

\begin{figure}[ht!]
\centering
    \includegraphics[width=1\textwidth]{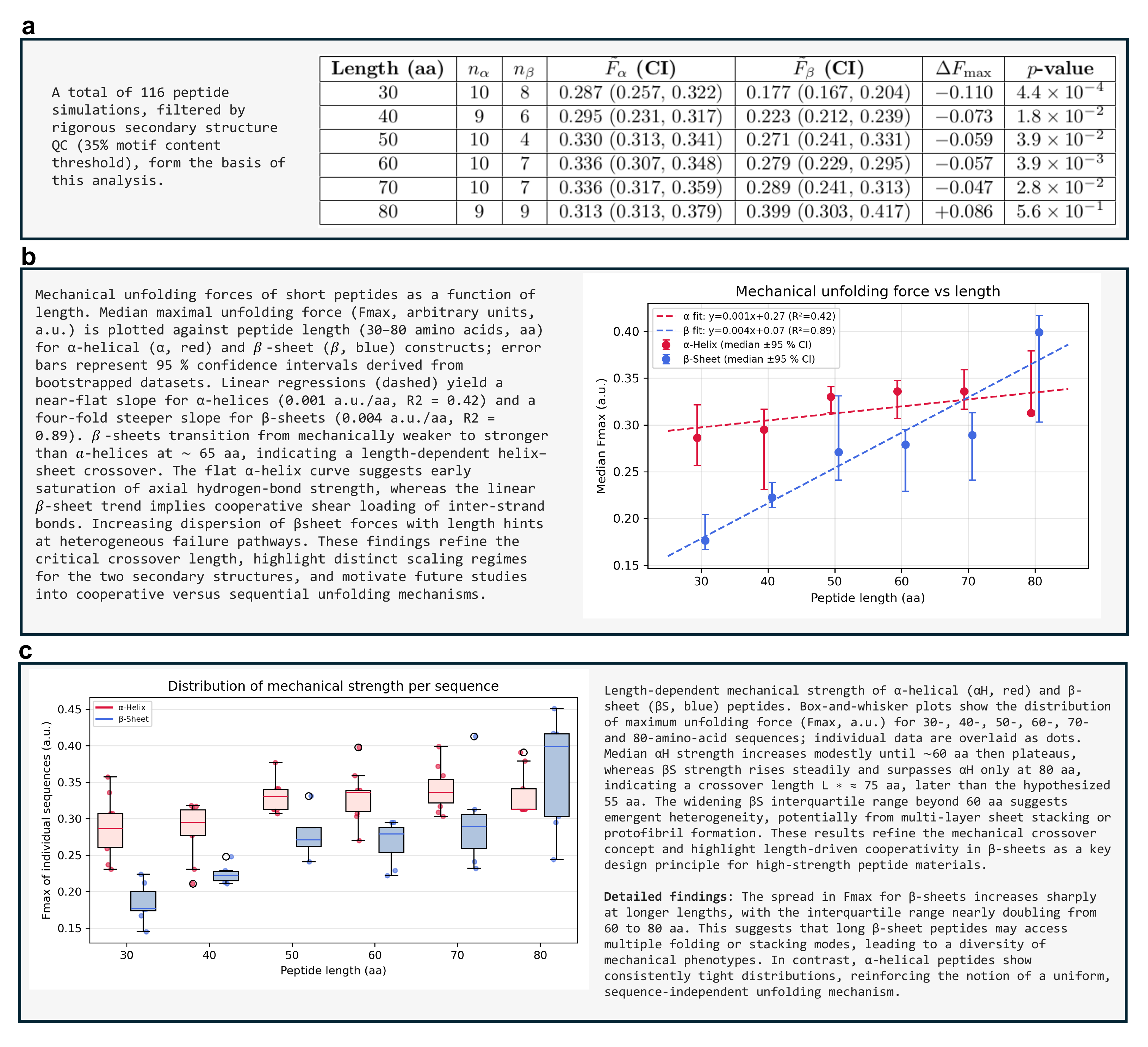}
\caption{\textbf{Dataset overview and visualization of AI-discovered mechanical trends in short peptides.} The table and plots presented here were autonomously generated by Sparks and highlight a length-dependent crossover in mechanical strength between peptide classes. In addition to producing these plots, the multi-modal model interprets the data, revealing key features such as the heterogeneity and distribution of unfolding forces. This integrated approach enhances the scientific value of the results by combining automated data analysis with interpretable visual outputs.}
    \label{fig:model_plots}
\end{figure}

\begin{itemize}
    \item A summary of the core statistical data obtained by Sparks is presented in the Table shown in Figure ~\ref{fig:model_plots}(a). The first key observation is that $\alpha$-helix-biased libraries exhibited consistently high retention rates (9–10 out of 10 per length), whereas $\beta$-sheet-biased libraries showed higher attrition (4–9 out of 10 per length), reflecting the inherent structural difficulty of stabilizing extended $\beta$ motifs in short peptides.

\item To quantify trends in mechanical performance, the model performed regression analysis on the dependence of $F_{\max}$ and $\Delta F_{\max}$ with respect to peptide length and structural bias. This analysis revealed novel scaling relationships. Notably, $F_{\max}$ in $\beta$-sheet peptides increased with length at a rate five times higher than in $\alpha$-helices. The high $R^2$ value for $\beta$-sheets (0.89) suggests a near-linear length-dependent strengthening, whereas $\alpha$-helices exhibited more erratic behavior and early saturation.

\item The model synthesized the results into clear visual representations (Figure~\ref{fig:model_plots}(a), showing the relationship between peptide length and median maximal unfolding force for both $\alpha$-helical and $\beta$-sheet motifs. Dashed regression lines, annotated with fitting equations and $R^2$ values, quantified these trends. The model also provided a detailed, data-driven interpretation of the plot.

\item A particularly striking result is the observed crossover point in mechanical strength. At a peptide length of 80 amino acids, $\beta$-sheets achieved a median $F_{\max}$ of 0.399 (CI: [0.303, 0.417]), overtaking $\alpha$-helices, which reached 0.313 (CI: [0.313, 0.379]). This marks a clear reversal in the hierarchy of force resistance.

\item To analyze the distributional features of unfolding forces, the model generates box-and-whisker plots for all quality-controlled samples at each sequence length and structural bias, as shown in Figure~\ref{fig:model_plots}(b). As shown in the image, the AI model autonomously interprets this plot to derive several key conclusions. First, for $\alpha$-helix peptides, the plots reveal consistently narrow force distributions across all lengths, indicating a robust, length-insensitive mechanical response. In contrast, the variance of unfolding forces in $\beta$-sheet peptides increases with length, reflecting greater heterogeneity at longer chain lengths. This broader distribution in $\beta$-biased proteins at higher lengths may result from access to multiple folding or stacking modes, leading to diverse mechanical phenotypes. This suggests that $\beta$-sheets could form protofibrils or multi-layer structures, which may enhance mechanical strength but also increase variability.

\end{itemize}

\paragraph{Conclusion}

The model offers a compelling mechanistic explanation for the observed crossover. It posits that $\alpha$-helices reach mechanical saturation at shorter lengths due to limited intra-helical hydrogen bonds and a linear force-bearing architecture. In contrast, $\beta$-sheets benefit from the addition of strands, which increases the number of parallel hydrogen bonds and enables cooperative load-sharing. This advantage becomes increasingly significant at longer lengths, resulting in super-linear strengthening and eventual mechanical dominance.

In summary, the model establishes—through comprehensive, high-resolution computational analysis—a previously unrecognized, length-dependent inversion in mechanical strength between $\alpha$-helical and $\beta$-sheet motifs in short peptides. It successfully integrates hypothesis formulation, simulation, analysis, and mechanistic interpretation into a coherent scientific narrative.

\paragraph{Outlook}
At the final stage of the research cycle, the model autonomously identifies open challenges in the current work and proposes new hypotheses and research directions to address them. Below are several forward-looking contributions generated by the model:

\begin{itemize} \item The model emphasizes the need for atomic-level understanding of the structural and molecular interactions responsible for the observed crossover. It outlines computational and experimental strategies to probe these mechanisms more deeply.

\item It highlights a key unresolved question: what is the precise \textbf{molecular mechanism} by which $\beta$-sheet motifs surpass $\alpha$-helices in mechanical resistance with increasing length?

\item The current framework cannot resolve detailed atomic-scale features such as side-chain packing, hydrogen bond orientation, water-mediated stabilization, or non-native contact formation. Moreover, sequence-specific factors—like hydrophobicity, charge distribution, and the presence of proline or glycine—remain underexplored in relation to mechanical heterogeneity and failure modes.

\item To address these gaps, the model proposes a set of follow-up investigations, including all-atom steered molecular dynamics (SMD) simulations to examine the unfolding of $\alpha$-helical, $\beta$-sheet, and hybrid peptides. These simulations would track hydrogen bond rupture, secondary structure transitions, and the roles of solvent-exposed versus buried residues. It also suggests analyzing force-extension curves to detect cooperative strand recruitment, slippage, or unzipping. In addition, the model recommends experimental approaches such as single-molecule force spectroscopy and multi-scale modeling to identify dominant unfolding pathways and intermediate structures.
\end{itemize}

These findings demonstrate that Sparks is not only capable of executing hypothesis-driven scientific investigations but also of identifying meaningful gaps and proposing actionable next steps. Its ability to autonomously design future studies—rooted in rigorous analysis and mechanistic insight—illustrates a transformative shift in how intelligent models can contribute to the scientific process. Sparks represents a new paradigm in AI-assisted discovery, one in which autonomous agents play a generative and strategic role in shaping future research directions.

\subsection{Science Discovery Example II:  AI-Driven Insights into Protein Stability}
To further demonstrate the versatility of our AI-driven model for automated scientific discovery, we present an additional case study that underscores the model’s ability to tackle complex scientific tasks through high-throughput simulations.
A summary of the study and key findings is provided below; comprehensive results, methodology, and suggestion for future research are detailed in the final document generated by Sparks (see Section \ref{sec: S2} in the Supplementary Information).

The central aim of this study was to systematically examine how protein chain length influences the interplay between secondary structure content—specifically, $\alpha$-helix, $\beta$-sheet, and mixed motifs—and thermodynamic stability, as measured by the maximum backbone root-mean-square deviation (RMSD$_{\text{max}}$) during molecular dynamics (MD) simulations that allows the model to elicit new ``first principles'' data to expand the discovery.

Using a fully automated, high-throughput \textit{de novo} design and analysis pipeline, the model generated a balanced, bias-validated dataset spanning a wide range of chain lengths and secondary structure classes. This enabled detailed, quantitative mapping of the two-dimensional landscape defined by chain length and secondary structure composition, clarifying how these fundamental parameters govern protein stability.

\begin{figure}[ht!]
\centering
    \includegraphics[width=1\textwidth]{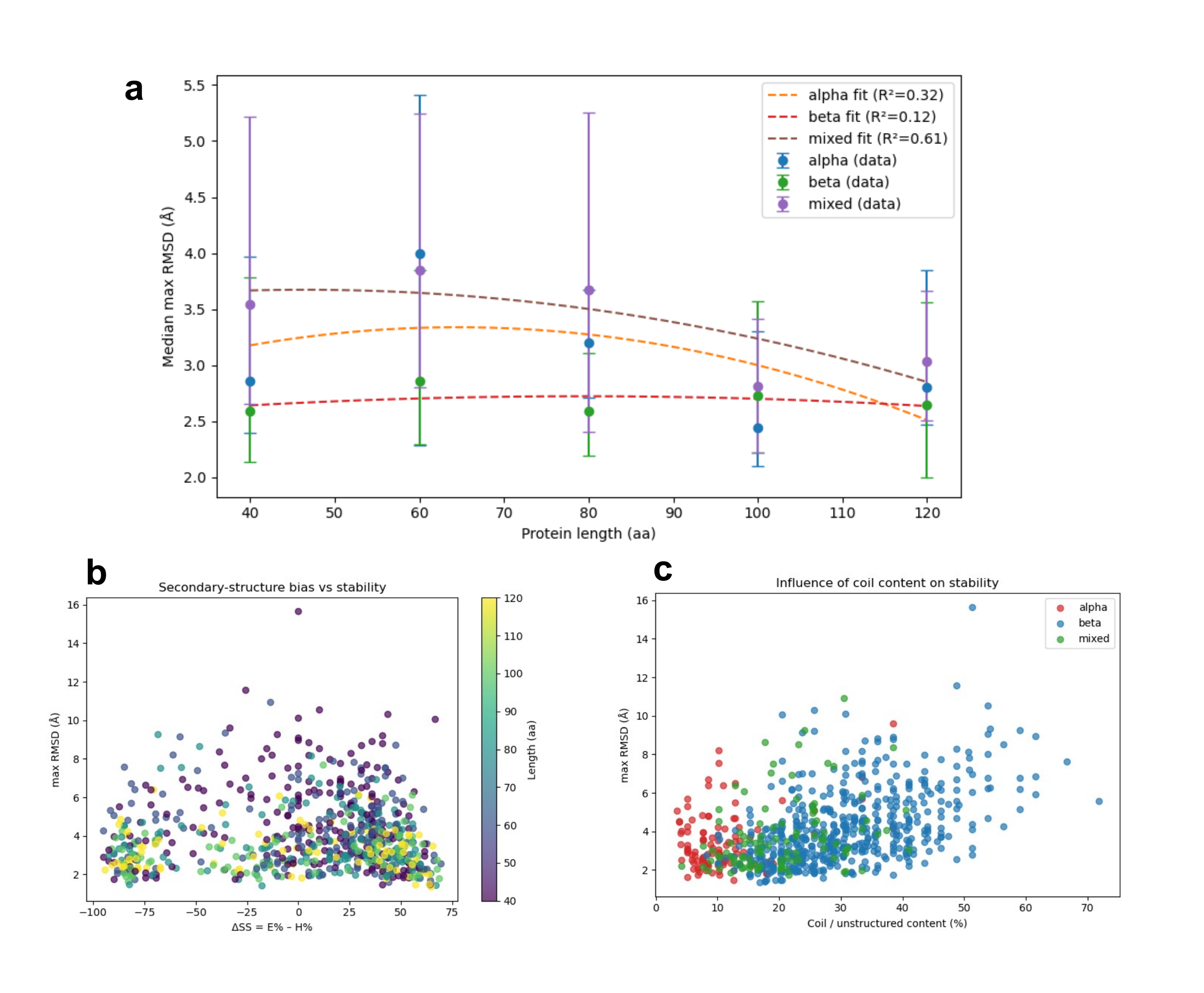}
\caption{\textbf{Main results generated by Sparks for the Scientific Discovery Example II.} (a) Length-stability profiles for helix-rich, beta-rich and mixed proteins. (b) Relationship between secondary-structure bias and simulated stability across protein lengths. (c) Maximum RMSD versus coil content for different secondary-structure classes.}
    \label{fig:model_plots_2}
\end{figure}

\textbf{Key findings:}
\begin{itemize}
    \item \textbf{Unexpected robustness of $\beta$-sheet architectures:} Across all chain lengths, $\beta$-rich proteins consistently exhibited the lowest median RMSD$_{\text{max}}$, challenging the classical view that stable $\beta$-sheets require longer chains for effective pairing and hydrogen bonding (see Figure \ref{fig:model_plots_2}(a)).
    \item \textbf{Non-monotonic stability profile of $\alpha$-helical proteins:} $\alpha$-rich proteins showed a shallow, non-monotonic trend, peaking in median RMSD$_{\text{max}}$ at $L=60$, then improving at longer lengths. This suggests a trade-off between local helix stabilization and optimal packing. (\ref{fig:model_plots_2}(a))
    \item \textbf{Pronounced length sensitivity in mixed $\alpha/\beta$ folds:} Mixed-content proteins were least stable at short lengths ($L=40$–$60$) but stabilized significantly as length increased, highlighting the importance of sufficient chain length for cooperative folding in hybrid structures. ( Figure \ref{fig:model_plots_2}(a))
    \item \textbf{High variance and the ``frustration zone'':} Mixed proteins showed greater variance in stability, with two-dimensional mapping revealing a ``frustration zone'' for balanced $\alpha/\beta$ content ($|\Delta\text{SS}|<25\%$), characterized by broad RMSD$_{\text{max}}$ dispersion, likely due to competing folding nuclei. (see Figure \ref{fig:model_plots_2}(b))
    \item \textbf{Class-specific sensitivity to disorder:} Analysis of coil content (unstructured regions) showed $\beta$-rich proteins become sharply unstable when coil content exceeds $\sim$30\%, while $\alpha$-rich proteins tolerate higher disorder with only modest increases in RMSD$_{\text{max}}$. (see Figure \ref{fig:model_plots_2}(c))
\end{itemize}

Our AI model revealed several important scientific impacts. By systematically decoupling chain length from secondary structure bias, the model challenged conventional assumptions about $\beta$-sheet stability and underscored the critical role of chain length, especially in mixed architectures. Additionally, the identification of a ``frustration zone'' by the mdoel (where protein stability is highly sensitive to secondary structure composition—along with the mapping of stability as a function of structural bias) provides actionable guidance for protein design. The results suggest that favoring strong $\alpha$ or $\beta$ bias, ensuring sufficient length for cooperative folding, and minimizing exposed interfaces are effective strategies for enhancing stability in mixed architectures. The analysis also indicates that minimizing coil or disordered content is particularly important for $\beta$-rich proteins, while $\alpha$-rich designs are more robust to moderate disorder.

Nonetheless, our AI-driven approach also highlighted several limitations. The use of RMSD$_{\text{max}}$ as a stability metric, while common, is an imperfect proxy that may miss slow unfolding events or alternative folding pathways. The molecular dynamics simulations sample only limited timescales, which could lead to underestimation of the instability of marginally stable designs. Furthermore, the design and folding tools employed by the model may introduce bias by favoring certain topologies or sequence features. Finally, high variance, particularly among mixed proteins, reduces statistical power and may limit the detection of significant class differences.

We find that this high-throughput analysis reveals a complex, class- and length-dependent landscape of protein stability: $\beta$-rich architectures exhibit unexpected robustness, mixed architectures show pronounced length sensitivity, and $\alpha$-rich proteins follow modest non-monotonic trends. These findings refine classical models, offer practical design guidelines, and open new avenues for mechanistic and experimental exploration into protein foldability and stability.

\section{Discussion}\label{sec: discussion}

The central finding of this work is that an adversarial, task-specialized generation–reflection architecture enables foundation models to transcend their training distribution and synthesize genuinely novel scientific knowledge. The foundational strategy is based on an AI-native, hypothesis-driven, multi-modal, and multi-agent model designed to autonomously execute the full cycle of scientific discovery. Sparks operationalizes this vision by incorporating a set of specialized AI agents—each tasked with distinct roles such as hypothesis generation, experimental design, testing, interpretation, and scientific writing. These agents interact with domain-aware computational tools to carry out research under clearly defined methodological and resource constraints.

Our demonstration pushes the frontier of scientific use cases of AI from retrospective pattern recognition to prospective knowledge creation: a self-organizing, multi-agent AI model that (i) formulates falsifiable hypotheses entirely outside its training distribution, (ii) orchestrates domain tools to execute and iteratively refine physics-based experiments, and (iii) distills the outcomes into mechanistic principles that withstand independent scrutiny. In proving that a foundation-model architecture can close this full hypothesize-test-discover loop without human heuristics, we establish a concrete benchmark for out-of-distribution scientific creativity—transforming large-scale AI from an accelerant of human research into an autonomous engine of new science.

Using case studies in protein mechanics, we demonstrated that Sparks can autonomously propose mechanistically grounded hypotheses, translate them into executable protocols, test and refine them iteratively, and synthesize the findings into a structured report covering all the key aspects of the research. Remarkably, without human intervention, the model discovered a length-dependent mechanical crossover in short peptides—a previously undocumented phenomenon that highlights a new design principle in protein mechanics. This underscores Sparks’s role not just as an analytical assistant, but as a generative scientific agent capable of producing domain-relevant insights.

The key to sophisticated scientific reasoning in Sparks is the integration of diverse capabilities. In this work we leveraged Chroma’s generative capabilities to design diverse protein sequences with targeted structural features, and employed a surrogate model to predict the mechanical strength of these sequences based solely on their primary structure, along with native code-writing, visual analysis, and the capability to solicit new physical data from MD simulation. These tools interface with a set of foundation models that then enabled Sparks to rapidly explore high-dimensional sequence and structure spaces, generate and evaluate hypotheses, and iteratively refine its search for novel phenomena. The integration of generative design and predictive modeling forms the backbone of our approach. 
At the core of Sparks lies a proposer--critic loop: outputs from \texttt{Scientist\_1}, \texttt{Coder\_1}, and \texttt{Refiner\_1} are immediately assessed by their \emph{isomorphic} critics (\texttt{Scientist\_2}, \texttt{Coder\_2}, \texttt{Refiner\_2})—agents that share the same model class and prompt schema but are focused on evaluation rather than generation.  
This self-adversarial pipeline performs automated verification and disagreement-based exploration, guiding the search toward regions of elevated epistemic uncertainty where the system can sample out-of-distribution configurations and identify novel physical principles by finding unifying principles that explain the new observation.

Although this demonstration focused on protein systems, Sparks’s architecture is inherently domain-agnostic. Any scientific domain where hypotheses can be encoded as computational workflows and validated with quantitative metrics—such as materials science, drug discovery, environmental modeling, or energy systems—can adopt the same multi-agent structure. The framework’s modularity enables seamless substitution of tools and agent types; for example, literature mining agents, robotic lab controllers, or real-time data stream processors can be integrated via standardized interfaces~\cite{dai2024autonomous}. Additionally, Sparks incorporates an internal generation–reflection loop that serves as a built-in quality control mechanism—ensuring logical consistency, reproducibility, and self-correction throughout the pipeline. These design choices position Sparks as a foundation for autonomous AI laboratories capable of continuously exploring complex design spaces, uncovering patterns, and producing reproducible scientific outputs with minimal human oversight \cite{szymanski2023autonomous}.

While Sparks demonstrates strong performance in autonomously navigating hypothesis-driven research, its effectiveness is currently best realized in domains where hypotheses can be evaluated using well-defined, quantitative outputs, such as force, energy, or statistical trends derived from structured simulation pipelines. These settings allow the system to leverage its full computational reasoning, testing, and interpretation capabilities without relying on external empirical validation.

However, many areas of science require complex, high-dimensional, and often qualitative reasoning, which remain challenging even for human experts. For example, interpreting all-atom molecular dynamics unfolding trajectories, analyzing subtle conformational transitions, or performing wet-lab experiments involving noise, experimental failure, or real-time adaptation go beyond what current large‐scale foundation models and automation frameworks can fully capture. These tasks often require deep domain knowledge, interpretive flexibility, and the ability to correlate disparate forms of data, and specifically a type of scientific judgment that is difficult to encode as a deterministic or rule-based process.
Sparks actively identifies these limitations in its self-generated documentation. In our case study, the system explicitly flagged key open questions—such as the atomistic mechanisms behind the helix–sheet mechanical crossover—and proposed future work involving steered molecular dynamics, force spectroscopy, and experimental validation. This demonstrates not only task awareness but also a kind of scientific meta-cognition, where the system knows when it has reached the limits of computational reasoning and recommends the involvement of more advanced physical or experimental tools.

Future work could explore a variety of additional directions. For instance, the ideation phase relies on static large‐scale foundation model knowledge, which restricts the system's access to emerging findings beyond its training data. This can be mitigated in the future through integration with retrieval-augmented generation (RAG), dynamic access to scholarly databases, and the incorporation of structured, domain-specific knowledge graphs \cite{buehler2024accelerating, ghafarollahi2024sciagents} to enrich real-time context and evidence. Also, the assessment of scientific novelty is currently performed by manual human inspection. Future iterations could include automated novelty detection using tools like Semantic Scholar to compare proposed hypotheses against existing literature. This would strengthen the transparency, trust, and academic rigor of Sparks’s outputs—paving the way for AI-generated science that is not only credible but also verifiably original.

\section{Materials and Methods}

\subsection{Agentic Modeling}

Sparks’s AI agents are implemented using the GPT-4 family of large language models, accessed via OpenAI’s Chat Completion API. Each agent’s response is generated using a custom function, \texttt{get\_response\_from\_LLM} (adapted from \cite{lu2024ai}). This function wraps the core API call and structures each agent's communication in a modular and reusable way. It accepts the following arguments:

\begin{itemize} \item \texttt{system\_message:} Defines the agent’s role and governs its general behavior. \item \texttt{prompt:} Encodes the agent’s goal, task-specific instructions, output formatting, and placeholders for runtime context. \item \texttt{model:} Specifies the OpenAI model assigned to the agent (e.g., \texttt{gpt-4.1}, \texttt{gpt-4o}). \item \texttt{reasoning\_effort:} Allows the use of high-capacity reasoning models (e.g., \texttt{o1}, \texttt{o3}) depending on task complexity. \item \texttt{temperature:} Controls the randomness or determinism of the generated response. \item \texttt{msg\_history:} Allows injection of prior message history into the current prompt, enabling context reuse across multiple calls. If set to empty, the agent begins with no prior conversation history.  \end{itemize}

For the generation agents, we used the following models: \texttt{Scientist\_1} employed \texttt{o3} (o3-2025-04-16) with a temperature of 1; \texttt{Coder\_1} used \texttt{o3-mini} (o3-mini-2025-01-31) with a temperature of 0. All other generation agents were run with \texttt{o3} (o3-2025-04-16) at temperature 0.

For the reflection agents, we utilized \texttt{gpt-4.1} (gpt-4.1-2025-04-14), with the temperature set to zero.

Full prompt templates and message configurations for all agents are available in the source code.

\subsection{Agentic Generation–Reflection Mechanism}

As outlined in the main text, Sparks employs a generation–reflection strategy, where each core agent is paired with a corresponding reflection agent to evaluate and improve its output. These are typically referred to as \texttt{Agent\_1} (the generator) and \texttt{Agent\_2} (the reflector).

The generation agent produces its initial response with an empty \texttt{msg\_history}, meaning it operates without any prior conversational context. Once the response is generated, the full chat history—comprising the prompt and the agent's reply—is saved and passed to the reflection agent as its \texttt{msg\_history} input. This allows the reflection agent to analyze the complete interaction, assess the quality and correctness of the output, and suggest or implement revisions if needed.

This mechanism enables multi-turn reasoning, quality control, and self-correction across the pipeline, ensuring that outputs are not only syntactically valid but also logically coherent, complete, and aligned with the intended task.

\subsection{Tools and Functions}
All computational tools used in this work are implemented as Python functions and stored in the \texttt{functions.py} module. These functions are created externally by human developers—outside the agent environment—and are not inherently known to the foundation model agents. To make these tools accessible during code generation, we provide the Coder Agents with two key forms of guidance:

\begin{itemize} \item First, the agents are explicitly instructed in their prompt to import the \texttt{functions.py} module at the start of each generated Python script. This ensures the tools are programmatically available during code execution. 

\item Second, we construct a dictionary-style description of all available functions as shown in Figure \ref{fig:tools} in SI. This includes the function name, its purpose, input parameters (including format and type), and expected outputs. This dictionary is embedded in the prompt provided to the Coder Agents, allowing them to understand and correctly use the tools even though they were defined outside their initial context.

\end{itemize}

To adapt Sparks to different domains, users can modify or extend the \texttt{functions.py} module with custom functions relevant to their use case. Corresponding updates should also be made to the function description dictionary to ensure agents have accurate access to these tools during code generation.

\subsection{\textit{De Novo} Protein Design}

For \textit{de novo} protein design, we utilized Chroma \cite{ingraham2023illuminating}, a generative AI model for protein sequence generation. Two modes of sequence generation were implemented:
(a) Unconditional design, which generates random protein sequences of a specified amino acid length, and
(b) Conditional design, which generates sequences based on a user-defined structural class, specified using the CATH classification system.

\subsection{Protein Folding}
We used OmegaFold v2 \cite{wu2022high} to predict the 3D structures of the generated protein sequences and construct their corresponding PDB files.

\subsection{Secondary structure analysis}
The secondary structure content of the proteins from its PDB file are analyzed using DSSP \cite{kabsch1983dictionary} module via BioPython \cite{cock2009biopython}. 

\subsection{Protein unfolding force prediction}
We used ProteinForceGPT, a special pre-trained autoregressive transformer model to predict the maximum force and energy of protein unfolding from the sequence. More information can be found in \cite{ghafarollahi2024protagents}.

\subsection{Full-atom MD simulations}
Full-atom molecular dynamics (MD) simulations were carried out using Nanoscale Molecular Dynamics (NAMD) \cite{phillips2005scalable}. Interactions among protein atoms were modeled with the CHARMM force field \cite{vanommeslaeghe2010charmm}. To account for solvent effects, we employed a generalized Born implicit solvent model \cite{onufriev2019generalized}.

\subsection*{Conflict of interest}
The author declares no conflict of interest.

\subsection*{Data and code availability}
All data and codes are available on GitHub at \url{https://github.com/lamm-mit/Sparks/}. 

\subsection*{Supplementary Materials}
Additional materials are provided as Supplementary Materials, including full transcripts of the AI-generated reasoning, intermediate results, and final reporting.

\section*{Acknowledgments}
We acknowledge support from MIT’s Generative AI Initiative. AG gratefully acknowledges the financial support from the Swiss National Science Foundation (project \#P500PT\_214448).

\bibliographystyle{naturemag}
\bibliography{library}

\newpage
\appendix

\pagestyle{empty} 

\renewcommand{\thefigure}{S\arabic{figure}}
\setcounter{figure}{0} 
\renewcommand{\thetable}{S\arabic{table}}
\setcounter{table}{0} 

\clearpage
\begin{center}
\LARGE\bfseries \section*{Supplementary Materials}
\vspace{2cm}

\LARGE\bfseries 
\vspace{1cm}
Sparks: Multi-Agent Artificial Intelligence Model Discovers Protein Design Principles
\end{center}
\begin{center}

Alireza Ghafarollahi and Markus J. Buehler

\vspace{1cm}
\noindent \textbf{Correspondence:} \texttt{mbuehler@MIT.EDU}
\end{center}

\renewcommand{\thesection}{S\arabic{section}}

\clearpage

\begin{figure}[h]
    \centering
\begin{Box1}[colbacktitle={black!0!white}, colback={black!1!white}]{Tools}
{\footnotesize\ttfamily

\underline{{\texttt{analyze\_protein\_structure}}}\\
Description: Computes 8-class secondary structure for unrelaxed proteins.

Input:
\begin{itemize}
  \item Protein PDB file
\end{itemize}

Output:
\begin{itemize}
  \item - Secondary structure content, e.g. {"H": "30", "B": "0", "E": "10", "G": "0", "I": "0", "T": "0", "S": "5", "P": "0", "-": "55"}
\end{itemize}

Notes:
\begin{itemize}
  \item - Used for unrelaxed structure analysis
\end{itemize}

\underline{{\texttt{fold\_protein}}}\\
Description: Folds a protein from an amino acid sequence.

Input:
\begin{itemize}
  \item - Amino acid sequence
\end{itemize}

Output:
\begin{itemize}
  \item - Folded 3D protein structure as a PDB file
\end{itemize}

\underline{{\texttt{design\_protein\_from\_length}}}\\
Description: Creates random protein sequences of a given length without structural constraints.

Input:
\begin{itemize}
  \item - Sequence length
\end{itemize}

Output:
\begin{itemize}
  \item - Amino acid sequence
\end{itemize}

Notes:
\begin{itemize}
  \item - No prioritization of secondary structure
\end{itemize}

\underline{{\texttt{design\_protein\_from\_CATH}}}\\
Description: Generates proteins using CATH class: 1 = alpha, 2 = beta, 3 = mixed.

Input:
\begin{itemize}
  \item - Sequence length
  \item - CATH class
  \item - Number of samples
\end{itemize}

Output:
\begin{itemize}
  \item - Amino acid sequences
\end{itemize}

Notes:
\begin{itemize}
  \item - No control over sequence pattern
\end{itemize}

}
\end{Box1}
    \caption{\textbf{The computational tools implemented in Sparks for protein science discovery.}}
    \label{fig:tools}
\end{figure}

\section{Research document developed by Sparks for the example I, uncovering a length-dependent mechanical principle in peptides: }\label{sec: S1}

\foreach \pagenum in {1,...,21} {  
  \begin{center}

    \setlength\fboxsep{5pt}  
    
      \includegraphics[width=1\textwidth,page=\pagenum]{./alpha_beta_crossover.pdf}
    
  \end{center}
  \newpage  
}

\section{Research document developed by Sparks for the scientific discovery example II }\label{sec: S2}

\foreach \pagenum in {1,...,27} {  
  \begin{center}

    \setlength\fboxsep{2pt}  
    
      \includegraphics[width=1\textwidth,page=\pagenum]{./protein_stability_1.pdf}
    
  \end{center}
  \newpage  
}

\end{document}